\documentstyle[uk-acl93]{article}

\newcommand{\nll}{$\cal N\!LL$}
\newcommand{\tdl}[0]{${\cal T\!D\!L}$}
\newcommand{\udine}[0]{${\cal U\!D}i\!{\cal N}\!e$}

\newcommand{\cosma}{{\sc cosma}}
\newcommand{\disco}{{\sc disco}}

\newcommand{\dfki}{DFKI}

\newcommand{\fegramed}{{\sc fegramed}}

\begin{document}

\title{\LARGE\bf \disco\ ---
\Large\bf An HPSG-based NLP System\\ and its Application for
Appointment Scheduling\\
\large\bf --- Project Note ---}
\author{Hans~Uszkoreit, Rolf~Backofen, Stephan~Busemann,
Abdel~Kader~Diagne,\\
\bf Elizabeth~A.~Hinkelman, Walter~Kasper,
Bernd~Kiefer, Hans-Ulrich~Krieger,\\\bf Klaus~Netter,
 G{\"u}nter~Neumann,
Stephan~Oepen, Stephen~P.~Spackman\\[0.5ex]
German Research Center for Artificial Intelligence (DFKI)  \\[-0.5ex]
Stuhlsatzenhausweg 3, 66123 Saarbr\"ucken, Germany  \\
{\tt $\langle${\em name}$\rangle$@dfki.uni-sb.de}}

\makesmalltitle
\begin{abstract}
{\small
The natural language system \disco\ is described. It combines

\begin{itemize}
 \setlength{\itemsep}{-2pt}
 \item  a powerful and flexible grammar development system;
 \item  linguistic competence for German including morphology,
        syntax and semantics;
 \item  new methods for linguistic performance modelling on the basis of
        high-level competence grammars;
 \item  new methods for modelling multi-agent dialogue competence;
 \item  an interesting sample application for appointment scheduling
        and calendar management.
\end{itemize}}
\end{abstract}

\section{Introduction}

We will describe results of a project in natural language research
carried out during the last four years at the German Research Center
for Artificial Intelligence in Saarbr{\"u}cken. All system building
took place during the last three years. The special approach of this
project is a combination of linguistically sound high-level grammatical
description and specialized methods for linguistic performance
modelling.

During the last decade it has become obvious to the majority of
researchers in our field that the linguistically designed elegant and
transparent grammars written in high-level representation languages
such as HPSG or LFG could not be employed for efficient and robust
processing in a straightforward way.

Many researchers have therefore resorted to well-known older methods
such as ATNs or other augmented finite-state methods, to statistical or
connectionist methods, or to combinations of these. Several projects
participating in the demanding ARPA competitions fall in this category.

Many others have decided to settle for a compromise between high-level
description and efficient processing by strongly constraining their
formalisms.
The resulting formalisms are usually much closer to PROLOG and do
not contain a powerful multiple-inheritance type system; e.g.\ the Core
Language Engine (CLE) of SRI Cambridge \cite{Alshawi:92b}, its
derivative the GEMINI system of  SRI International in Menlo Park, the
LKP of SIEMENS in Munich \cite{Block:94}.
As a consequence of their design philosophy, these systems usually do
not feature a powerful development platform.

Only a few groups have continued to work in high-level formalisms
driven by the expectation that better processing methods for these
formalisms can be developed.
Our work belongs in this category. However, we have decided to allow
for combinations of high-level grammatical description and low-level
processing methods by strictly distinguishing between a general linguistic
competence model and very application-specific performance models.

Our decision was based on some fundamental criteria:

\begin{itemize}
  \setlength{\itemsep}{0pt}
  \item  It is necessary to build general, reusable competence
         systems that can be the basis for different applications, since the
         development of linguistic competence is too costly to redo
         for each new application.

  \item  In the foreseeable future, NL applications will have very
         limited linguistic competence, where the limitations depend on
         the task to be performed. They determine coverage and depth of
         analysis.
  \item The general competence system as such will not be used for
         individual applications because each application type imposes
         specific requirements on the performance model. Depending on
         the task there are quite different constraints on robustness,
         accuracy, and processing speed.
\end{itemize}

On the basis of these assumptions, we took a rather uncompromising
stand.  We decided to utilize the most suitable and most advanced
methods for the development of linguistic competence. Our development
platform is based on a powerful typed feature unification formalism, and
the grammar follows the HPSG theory.  These choices were made since we
wanted on the one hand to facilitate the difficult and time-consuming
process of grammar development, and on the other to save our grammar
from the fate of several older large-coverage grammars which cannot be
reused or extended today because of their idiosyncratic representations.

Since research on systems with multiple cooperating agents constitutes
one of the focal areas of our institute, we tried to develop the system
in such a way that it would support dialogue among such agents. At the
same time, we undertook serious efforts in research on methods that
would allow us to derive adequate performance models from the core
competence system.

We also built a sample application (\cosma) for appointment scheduling and
management based on the competence model, in order to test  the grammar of
German, methods for dialogue modelling, and certain new methods for deriving
a performance model from the competence system.

In the remainder of this paper we will present an overview of the following
components and methods:
\begin{itemize}
  \setlength{\itemsep}{0pt}
  \item  development  platform including shell, formalism, morphology,
         parser, generator, semantics;
  \item  German competence including morphology, syntax, semantics;
  \item  methods for providing multi-agent dialogue competence;
  \item  methods for linguistic performance modelling;
  \item  the NL functionality of the sample application \cosma.
\end{itemize}

Some individual components and methods have been described in
more detail in previous publications. However, this paper is the
first attempt 
to present an overview of  the integrated system
and to describe its parts from the perspective of our overall
research strategy.


\section{Formalism}

For the grammar, the lexicon and parts of the morphology a powerful typed
unification formalism \tdl\ (Type Description Language) has been
developed.  Reasoning is performed by two specialized inference
engines, viz.\ the \tdl\ type engine and the feature constraint-solver
\udine.
The modules are connected via a flexible interface to allow for mutual
control.

\paragraph{Type System} \tdl\ is a powerful typed feature-based
language and inference system, specifically suited for highly lexicalized
grammars \cite{Krieger&Schaefer:93} (in this volume).
Type definitions in \tdl\ consist of type constraints and feature
constraints over the standard boolean connectives $\wedge$, $\vee$, and
$\neg$.
The operators are generalized in that they can connect feature
descriptions, coreference tags (logical variables) and types.
\tdl\ distinguishes between {\em avm\/} types (open-world reasoning),
{\em sort\/} types (closed-world reasoning), {\em built-in\/} types, and
atoms.  Recursive types are explicitly allowed and handled by a sophisticated
lazy type expansion mechanism.

\tdl\ allows the definition of {\em partitions\/} and the declaration of
sets of types as {\em incompatible\/}, meaning that the conjunction of them
yields $\bot$.
Working with partially as well as with fully expanded types is possible
through the use of a sophisticated type expansion mechanism, both at
definition and at run time.
\tdl\ is fully incremental in that it allows the redefinition of types
and the use of undefined types.
\tdl\ allows a grammarian to define and use parameterized templates
(macros).  Input given to \tdl\ is parsed by a LALR(1) parser to
allow for an intuitive, high-level input syntax.

Efficient reasoning in the system is accomplished through specialized
modules:\ (i) bit vector encoding of the type subsumption hierarchy;
(ii) fast symbolic simplification for complex type expressions;
(iii) memoization to cache precomputed results;
and (iv) type expansion to make constraints explicit, to determine
the global satisfiability of a description, and to work with partially
expanded types during processing.

\paragraph{Constraint Solver}

\udine\ is a feature constraint solver capable of dealing with
distributed disjunctions over arbitrary structures, negative
coreferences, full negation, and functional and relational constraints.
It is the first (and to our knowledge the only) implemented feature
constraint solver that {\em integrates\/} both full negation and
distributed disjunctions~\cite{Bac:Wey:94}.
\udine\ does not use distributed disjunction only as a tool for efficient
processing. It also forms part of the input syntax, which allows for
very compact representation of the input data. In contrast with other
systems using distributed disjunctions, we do not restrict disjunctions
to length two, thus reducing the size of the feature structure
representations massively.

The functionality of \udine\ is completed by several auxiliary
functions. It is possible to remove inconsistent
alternatives, to simplify structures, to extract subterms or to evaluate
functional constraints.  One can also construct disjunctive normal form if
desired.

\paragraph{Semantic Representation}
A specialized meaning representation formalism, \nll, developed at
Hewlett Packard \cite{Lau:Ner:91}, is used for semantic reasoning and
as a flexible interface to various application systems.
\nll\ is a linguistically motivated extension of sorted first-order
predicate logic, integrating also concepts from Situation Semantics and
DRT.
It provides a large range of representational mechanisms for natural
language phenomena.


\section{Linguistic Resources}

The core of the linguistic resources consists of a two-level
morphology with feature constraints, an HPSG oriented grammar of
German with integrated syntax and semantics, and a module for
surface speech act recognition, all implemented in \tdl.

\paragraph{Morphology}
The component X2MorF, analyzing and generating word
forms, is based on a two-level morphology which is extended by a
word-formation grammar (described in \tdl) for handling the
concatenative parts of morphosyntax \cite{Trost:90}.

\paragraph{Grammar}
The style of the grammar closely follows the spirit of HPSG,
but also incorporates insights from other grammar frameworks (e.g.\
categorial grammar) and further extensions to the theory \cite{Net:93}.

The grammar distinguishes various types of linguistic objects, such as
lexical entries, phrase structure schemata, lexical rules, multi-word
lexemes etc., all of which are specified as typed feature structures.
Lexical rules are defined as unary rules and applied at run-time.
Multi-word lexemes are complex lexemes with a non-compositional
semantics, such as fixed idiomatic expressions.
HPSG principles and constraints are represented by inheritance links in
the type lattice.
The grammar covers a fair number of the standard constructions of
German, and exhibits a more detailed coverage in some specific
application oriented areas.

\paragraph{Semantics}
Feature structure descriptions of the semantic contribution of
linguistic items are represented in \tdl\ and are fully integrated into
the grammar.
Additionally, the \tdl\ type system is used to encode and check sortal
constraints as they occur in selectional restrictions.
For further processing such as scope normalization and anaphora resolution,
inferences and application dependent interpretation, the (initial)
\tdl\ semantic descriptions are translated into \nll\ formulae.

\paragraph{Speech Act Recognition and Dialogue}
The grammar provides a typed interface to a speech act recognition
module based on HPSG representations of utterances.  The assignments of
illocutionary force take into account syntactic features, a marking of
performative verbs and assignments of fixed illocutionary force to
relevant idiomatic expressions.

Recently inference-based dialogue facilities using a quasi-modal logic
for multiagent belief and goal attribution \cite{Hin:Spa:92} have been
added to the system.
Incoming surface speech act structures are subjected to anaphora and
reference resolution, translated into a frame-based action
representation, and disambiguated using inferential context.
The effects, including communicated beliefs and goals, of the first
acceptable speech act interpretation are then asserted.

\section{Processing components}

Parser and generator provide the basic processing functionality needed
for grammar development and sample applications.
In addition to the separate modules for parsing and generation, we also
experiment with a uniform reversible processing module based on
generalized Earley deduction.

\paragraph{Parser}
The parser is a bidirectional bottom-up chart parser which operates on a
context-free backbone implicitly contained in the grammar \cite{Kie:Sch:94}.
The parser can be parameterized according to various processing
strategies (e.g. breadth first, preference of certain rules etc.).
Moreover, it is possible to specify the processing order for the
daughters of individual rules.
An elaborate statistics component supports the grammar developer in
tuning these control strategies.

In addition, the parser provides the facility to filter out useless
tasks, i.e.\ tasks where a rule application can be predicted to fail by
a cheaper mechanism than unification.
There is a facility to precompute a filter automatically by determining
the possible and impossible combinations of rules; some additional
filtering information is hand-coded.

The parser is implemented in an object-oriented manner to allow for
different parser classes using different constraint solving mechanisms or
different parser instances using different parsing strategies in the same
system.
With differing parameter settings instances of the parser module are
used in the X2MorF and surface speech act recognition modules as well.

\paragraph{Generator}
Surface generation in \disco\ is performed with the SeReal (Sentence
Realization) system \cite{Busemann:93d}, which is based on the
semantic-head-driven algorithm by Shieber et al.
SeReal takes a \tdl\ semantic sentence representation as its input and
can deliver all derivations for the input admitted by the grammar.
Efficient lexical access is achieved by having the lexicon indexed
according to semantic predicates.
Each index is associated with a small set of lemmata containing the
semantic predicate.
Using the same indexing scheme at run-time for lexical access allows us
to restrict unification tests to a few lexical items.
Subsumption-based methods for lexical access were considered too
expensive for dealing with distributed disjunctions.
The grammar used for generation is the same as the one used for parsing
except for some compilation steps performed by SeReal that, among other
things, introduce suitable information wherever `semantically empty'
items are referred to.
Rule application is restricted by rule accessibility tables which are
computed off-line.

\section{Performance Modelling}

In our search for methods that get us from the transparent and extensible
competence grammar to efficient and robust performance systems
we have been following several leads in parallel.
We assume that methods for compilation, control and learning need to be
investigated.
The best combination of these methods will depend on the specific
application.
In the following some initial results of our efforts are summarized.

\paragraph{Acquisition of Sublanguages by EBL}
It is a matter of common experience that different domains make
different demands on the grammar.
This observation has given rise to the notion of sublanguage;
efficient processing is achieved by the exploitation of
restricted language use in well specified domains.

In the \disco\ system we have integrated such an approach based on
Explanation-Based Learning (EBL) \cite{Neumann:94}.
The idea is to generalize the derivations of training instances
created by normal parsing automatically and to use these generalized
derivations (called templates) in the run-time mode of the
system. If a template can be instantiated for a new input,
no further grammatical analysis is necessary.
The approach is not restricted to the sentential level but can also
be applied to arbitrary subsentential phrases, allowing it to
interleave with normal processing.

\paragraph{Intelligent Backtracking in Processing Disjunctions}
In \cite{Uszkoreit:90e} a method is outlined for controlling the order
in which conjuncts and disjuncts are to be processed.
The ordering of disjuncts is useful when the system is supposed to find
only the best result(s), which is the case for any
reasonably practical NL application.
An extension of \udine\ has been implemented that exploits distributed
disjunctions for preference-based backtracking.

\paragraph{Compilation of HPSG into Lexicalized TAG}
\cite{Kie:Net:Vij:94} describes an approach for compiling HPSG into
lexicalized feature-based TAG.
Besides our hope to achieve more efficient processing, we want to gain
a better understanding of the correlation between HPSG and TAG.
The compilation algorithm has been implemented and covers almost all
constructions contained in our HPSG grammar.

\section{Environment}

The {\sc disco development shell} serves as the basic architectural
platform for the integration of natural language components in the
\disco\ core system, as well as for the \cosma\ application system
\cite{neumann:93c}.
Following an {\em object oriented\/} architectural model we followed a
{\it two-step approach}, where in the first step the architecture is
developed independently of specific components to be used and of a
particular flow of  control.
In the second phase the resulting `frame system' is instantiated by the
integration of existing components and by defining the
particular flow of control between these components.
Using an object-oriented design together with multiple inheritance has
been shown fruitful for the system's  modifiability, extensibility
and incremental usability.

Several editing and visualization tools greatly facilitate the work of
the grammar developer.
The most prominent of them, \fegramed, provides the user with a fully
interactive feature editor and viewer.
There are many possibilities to customize the view onto a feature
structure, such as hiding certain features or parts of a structure,
specifying the feature order and many more.
The large feature structures emerging in the process of constraint
based formalisms make such a tool absolutely indispensable for grammar
debugging.
Main goals of the development of \fegramed\ were high portability and
interfacing to different systems.
Written in ANSI-C, it exists in Macintosh and OSF/Motif versions and
is already used at several external sites.

There exists a graphical chart display with mouse-sensitive chart nodes
and edges directly linked to the feature viewer, thus making debugging
much simpler.
It also provides a view of the running parser and enables you to
inspect the effects of the chosen parsing strategy visually.
A browser for the \tdl\ type system permits navigation through a
type lattice and is coupled with the feature editor.
There are other tools as well, e.g., a \tdl2\LaTeX utility, an {\sc Emacs}
\tdl\ mode, global switches which affect the behaviour of
the whole system etc.

The diagnostics tool (DiTo) \cite{Ner:Net:Dia:93} containing close to
1500 annotated diagnostic sentences of German facilitates consistency
maintenance and measuring of competence.
The tool has been ported to several sites that participate in
extending the test-sentence database.

\section{Putting it to the Test}

\paragraph{Cooperative Schedule Management}
In building the \cosma\ prototype the \disco\ core system has
been successfully integrated into an application domain with both
scientific interest and practical plausibility, viz.\ multi-agent
appointment scheduling (see Figure~\ref{cosma}).
Understanding and sending messages in natural language is crucial for
this application since it cannot be expected that all participants will
have a \cosma\ system.
The use of natural language also makes it easier for the owner of the
system to monitor the progress of an appointment scheduling process.
Each \cosma\ instance functions as a personal secretarial assistant
providing the following services: (i) storage and organization of a
personal appointment date-book; (ii) graphical display and manipulation
of appointment data; and (iii) natural language understanding and
generation in communication with other agents via electronic mail.
The current scheduling functionality includes the arrangement of
multi-participant meetings (possibly with vague or underspecified
details) as well as the modification and cancellation of appointments
that are under arrangement or have already been committed to.

Accordingly, the current \cosma\ architecture has three major
components: a prototype appointment planner (developed by the \dfki\
project AKA-MOD) that keeps the calendar database, provides temporal
resolution and drives the communication with other agents; a graphical
user interface (developed inside the \disco\ project) monitoring the
calendar state and supporting the mouse- and menu-driven arrangement of
new appointments and, finally, the \disco\ core system (enriched with a
set of application specific modules) that provides the natural language
and linguistic dialogue capabilities.

\begin{figure}
  \begin{center}
    \input{coling-figure}
  \end{center}
  \caption{Rough sketch of the \disco\ system in its application to the \cosma\
           scenario.  The entire \cosma\ prototype has been built on top of the
	   {\sc disco development shell\/} as a monotonic extension to the core
system.}
  \label{cosma}
\end{figure}

\paragraph{Interface to the Core Engine}
The communication between the \disco\ system and the
appointment planner is modelled in a restricted appointment task
interface language and roughly meets the internal representation of the
appointment planner.
To connect the two components, \disco\ is enriched with a
dedicated interface module that translates between the \disco\ internal
semantics representation language \nll\ and the appointment planner
representation.  The translation process (maintaining the substantial
difference in expressive power between \nll\ and the restricted planner
language) builds on ideas from current compiler technology with a
limited set of domain- and application-specific inference rules
\cite{Ner:Lau:Dia:93}.

On its opposite end \disco\ is hooked up to plain electronic mail
facilities through a general purpose e-mail interface that allows it to
receive and send e-mail (and in case of processing failures to
`respool' messages to the user mailbox).

\section{Discussion and Outlook}

We have presented an overview of the \disco\ system that illustrates our
general research strategy.
The system is implemented in Common Lisp and runs on Sun and HP
workstations.
Some tools and peripheral components are coded in C.
A port of the system to another hardware platform (Apple Macintosh) is
currently being carried out.
The total size of the system is about 100,000 lines of code.
Parts of the system were demonstrated at several conferences, at trade
fairs and on other occasions.

The work will be continued in two smaller projects.
First of all we plan to extend the system's linguistic competence of
German.
The diagnostic tool DiTo will be expanded in joint work with other
groups to provide an instrument for measuring progress and for
comparing grammars. We will also continue work on building up dialogue
competence.
The application demonstrator will be further developed in collaboration
with other projects at the DFKI.

In the area of performance modelling, we will continue exploring
different approaches for control, compilation, and competence-based
learning in parallel.
At this point nobody can really foresee which strategy or combination
of strategies will yield the best practical results.
As we pointed out in the introduction, different application types will
require different performance models.
High priority is given to the integration of statistical methods in all
pursued approaches, since in contrast to competence modelling, statistical
data are essential for developing adequate performance models.

\section*{Acknowledgements}

We acknowledge the invaluable input of our former colleagues, viz., John
Nerbonne who substantially contributed to the design and success of the
\disco\ project, Harald Trost and Jan Alexandersson.

As in many academic environments, major parts of the daily system building
have been carried out by our wonderful research assistants, especially
Sabine Buchholz, Stephan Diehl, Thomas Fettig, Stefan Haas, Judith Klein,
Karsten Konrad, Ingo Neis, Hannes Pirker, Ulrich Sch\"afer, Oliver Scherf,
J\"org Steffen, and Christoph Weyers.

This work has been supported by research grant ITW 9002 0 from the German
Bundesministerium f{\"u}r Forschung und Technologie to the \disco\ project.


\small
\bibliographystyle{plain}
\bibliography{coli-brand-new}


\end{document}